\documentclass[11pt]{article}

\usepackage[final]{acl}

\usepackage[normalem]{ulem}
\useunder{\uline}{\ul}{}
\usepackage{booktabs}       
\usepackage{amsfonts}       
\usepackage{nicefrac}       
\usepackage{microtype}      
\usepackage{xcolor}         
\usepackage{graphicx}
\usepackage{amsmath}
\usepackage{amssymb}
\usepackage{array}
\usepackage{multirow}
\usepackage{color}
\usepackage{colortbl}
\usepackage{framed}
\usepackage{bm}
\usepackage{xspace}
\usepackage{enumitem}
\usepackage{xparse}
\usepackage{algorithm}
\usepackage{listings}
\usepackage{url}
\usepackage{mathtools}
\usepackage{pifont}
\usepackage{mathtools}
\usepackage{lipsum}
\usepackage{adjustbox}
\usepackage{stfloats}
\usepackage{listings}
\usepackage{makecell}

\newcommand{\cmark}{\ding{51}}%
\newcommand{\xmark}{\ding{55}}%

\newcommand{\ours}{IBA\xspace}

\usepackage{times}
\usepackage{latexsym}

\usepackage[T1]{fontenc}

\usepackage[utf8]{inputenc}

\usepackage{microtype}

\usepackage{inconsolata}

\usepackage{graphicx}

\newcommand{\eg}{\emph{e.g.,}\xspace}
\newcommand{\ie}{\emph{i.e.,}\xspace}

\title{Ground Then Rank: Revisiting Knowledge-Based VQA with Training-Free Entity Identification}

\author{
 \textbf{Qian Ma\textsuperscript{1} \thanks{This work was initiated and done while the first author was an intern at AT\&T CDO. Qiong Wu and Yao Ma are co-corresponding authors.}},
 \textbf{Qiong Wu\textsuperscript{2}},
 \textbf{Zhengyi Zhou\textsuperscript{2}},
 \textbf{Yao Ma\textsuperscript{1}}
\\
 \textsuperscript{1} Rensselaer Polytechnic Institute
 \\
 \textsuperscript{2} AT\&T Chief Data Office
\\
 \small{
   \textbf{ \{maq5,may13\}@rpi.edu, \{qw6547,zz547k\}@att.com }
 }
}

\begin{document}
\maketitle

\begin{abstract}
Knowledge-Based Visual Question Answering (KB-VQA) requires grounding visual queries to external knowledge beyond directly observable content in images.
While recent multi modal large language models (MLLMs) show strong perceptual abilities, they struggle on KB-VQA tasks requiring groundings from both fine-grained entity and evidence levels.
Most existing multi-modal retrieval augmented generation (MM-RAG) methods tightly couple entity discrimination and section-level evidence ranking into a single re-ranking stage, leading to high cost and limited generalization.
In this work, we revisit existing MM-RAG solutions from a workflow perspective and argue both entity-level and fact-level groundings are key bottlenecks.
We observe that although MLLMs often fail under open-ended entity naming, they can better identify the correct entity when selecting from a small set of candidate names.
Based on this insight, we propose a simple and training-free \emph{identify-before-answer~\ours} framework that decouples entity identification from section-level re-ranking.
Our approach prompts an MLLM to select high-confidence entities using only candidate names, followed by an off-the-shelf textual re-ranker for evidence selection.
Experiments on Encyclopedic-VQA and InfoSeek show that our method consistently outperforms fine-tuned multi-modal re-ranking baselines while reducing training and inference complexity.
Additional analyses reveal that the improvements arise not only from better entity identification, but also from selecting more informative evidence once correct entity is fixed.
Our implementation is made public to ease reproducibility~{\url{https://github.com/VAN-QIAN/ACL26-IBA/}}.
\end{abstract}    
\section{Introduction}\label{sec:intro}

Knowledge-Based Visual Question Answering (KB-VQA) extends standard VQA by requiring external world knowledge beyond what is directly observable in the image~\cite{deng2025comprehensive,kim2025visual}. 
While recent multi-modal large language models (MLLMs) achieve strong performance on perception-driven VQA, KB-VQA queries often hinge on identifying the correct real-world entity and grounding fine-grained factual information that cannot be inferred from pixels alone~\cite{EVQA,InfoSeek}. 
This reliance on entity-level knowledge makes KB-VQA a challenging benchmark for multi-modal intelligence.

Modern MLLMs~\cite{liu2023visual,liu2024improved,bai2025qwen2} have demonstrated remarkable progress on general VQA tasks, yet they remain unreliable on KB-VQA where relevant knowledge is sparse, long-tailed, and difficult to encode in model parameters~\cite{kuang2025natural,deng2025comprehensive}. 
As a result, recent systems predominantly adopt a multi-modal retrieval-augmented generation (MM-RAG) paradigm~\cite{chen2022murag,yu2025mramg}, which first retrieves a set of potentially relevant knowledge entries (e.g., Wikipedia articles) and then re-ranks textual sections to support answer generation, as illustrated in Figure~\ref{fig:KB-VQA-workflow}.

Despite their success, we argue that existing MM-RAG methods suffer from a fundamental limitation in their workflow. 
Producing a correct and verifiable answer requires grounding at two distinct levels: 
(i) \emph{entity-level grounding}, ensuring that the retrieved context refers to the correct entity depicted in the image, and 
(ii) \emph{section-level grounding}, locating the passage within that entity's article that is relevant to the question. 
However, most existing approaches~\cite{yan2024echosight,ReflectiVA,CoRe-MMRAG} implicitly couple these two challenging tasks into a single re-ranking step over all candidate sections.
This entangled formulation forces a single scoring function to simultaneously discriminate between entities and rank textual evidence, often leading to textually relevant but entity-incorrect contexts, or correct entities paired with irrelevant sections.

In contrast, humans naturally decouple these two steps when solving KB-VQA problems.
After an initial retrieval or recall of plausible candidates, people typically first identify or narrow down the entity depicted in the image, and only then examine a small number of relevant articles to locate supporting evidence.
This decomposition reduces distractors and simplifies subsequent reasoning, suggesting a more principled paradigm.

Motivated by this observation, we revisit the role of MLLMs in entity identification.
While directly naming an entity from an image remains challenging for current models~\cite{caron2024generative}, we make a surprising empirical observation: simply providing candidate entity names enables MLLMs to identify the correct entity with much higher accuracy.
This suggests that MLLMs often possess incomplete yet usable entity knowledge that is difficult to exploit under open-ended generation but can be effectively activated when the task is reframed as a constrained discrimination problem.
This behavior bears resemblance to a tip-of-the-tongue-like effect~\cite{brown1966tip}, which we use only as an intuitive analogy.
ToT describes a situation that human experts may also encounter that they have the expertise to the entity but can't directly recall the name from scratch. But once several plausible names (\eg ``Lapsana communis'' and ``Crepis tectorum'' in Figure~\ref{fig:KB-VQA-workflow}) are presented, they can reason from visual cues with their expertise to select the correct one.

Based on this insight, we propose a simple yet effective \emph{\ours} framework for KB-VQA.
Our approach explicitly inserts a lightweight entity identification step into the MM-RAG workflow. 
After initial retrieval, the MLLM scores candidate entities using only their names, retains a small subset of high-confidence entities, and then applies a pre-trained standard textual re-ranker to select supporting sections within this narrowed scope.
This training-free design decouples entity recognition from evidence selection, improving both accuracy and efficiency.

Our contributions are summarized as follows:
\begin{itemize}[leftmargin=*]
    \item To the best of our knowledge, we are the first to report the `tip-of-the-tongue' phenomenon in MLLMs for KB-VQA, where providing candidate entity names significantly amplifies the model's reasoning capability to better identify the entity in the query image.
    \item Based on this finding, we propose a simple yet effective framework that integrates an explicit identification step into existing MM-RAG workflows, enhancing answer accuracy and computational efficiency without additional fine-tuning or task-specific training.
    \item We validate our approach on two mainstream KB-VQA benchmarks, achieving new state-of-the-art results while improving efficiency compared to existing MM-RAG systems.
\end{itemize}
\section{Related Works}

\subsection{MLLMs for KB-VQA}
Multimodal large language models (MLLMs) extend text-only LLMs with visual encoders and cross-modal alignment mechanisms, enabling joint reasoning over images and text.
Recent models such as LLaVA~\cite{liu2023visual,liu2024improved} and Qwen-VL~\cite{bai2025qwen2} achieve strong performance on perception-driven VQA benchmarks, where answers can be inferred directly from visual content or broadly learned parametric knowledge.

However, emerging evaluations~\cite{li2024cognitive,tan2025vision} consistently show that even state-of-the-art MLLMs underperform on knowledge-based VQA (KB-VQA) tasks that require fine-grained, entity-centric, or long-tail encyclopedic knowledge.
This limitation motivates augmenting MLLMs with external knowledge sources to support explicit grounding and reasoning beyond their parametric capacity

\subsection{KB-VQA and MM-RAG based solutions }

KB-VQA benchmarks extend conventional VQA by requiring external knowledge not contained in the image alone, such as entity attributes or encyclopedic facts.
Early datasets such as OK-VQA~\cite{marino2019ok} and A-OKVQA~\cite{schwenk2022okvqa} emphasize commonsense or general knowledge, which increasingly falls within the training scope of large-scale MLLMs.

More recent benchmarks, including Encyclopedic-VQA (E-VQA)\cite{EVQA} and InfoSeek\cite{InfoSeek}, raise the difficulty by requiring explicit grounding to fine-grained Wikipedia entities and supporting sections.
To address these challenges, many methods adopt multimodal retrieval-augmented generation (MM-RAG), typically consisting of a retriever, a re-ranking stage, and an answer generator.
Representative approaches such as EchoSight~\cite{yan2024echosight}, ReflectiVA~\cite{ReflectiVA}, and CoRe-MMRAG~\cite{CoRe-MMRAG} differ in how relevance is assessed, ranging from explicitly trained multimodal re-rankers to relevance implicitly learned through fine-tuning.
Despite their success, these methods generally couple entity discrimination and section selection into a single re-ranking process, which can be costly to train and sensitive to data availability.

\subsection{Positioning of our work}

In contrast to prior MM-RAG methods, our work revisits the KB-VQA pipeline from a workflow perspective.
Rather than entangling entity identification and section-level evidence selection within a single re-ranking module, we explicitly decouple these two stages.
By introducing a lightweight identification step before section re-ranking, our approach isolates the entity-level grounding problem and leverages off-the-shelf components without requiring task-specific re-ranker training or MLLM fine-tuning.
This design differs fundamentally from prior approaches that rely on learned multimodal relevance functions, and enables more interpretable and transferable KB-VQA pipelines across datasets with varying knowledge distributions.
\section{Methodology}

In this section, we present our training-free \ours (Identify Before Answer) framework, which explicitly decouples entity-level identification from section-level evidence selection.
We first introduce the overall workflow and then describe each component in detail, including problem formulation, initial retrieval, identify-before-re-rank, and answer generation.

\begin{figure*}[!htbp]
    \centering
    \includegraphics[width=\linewidth]{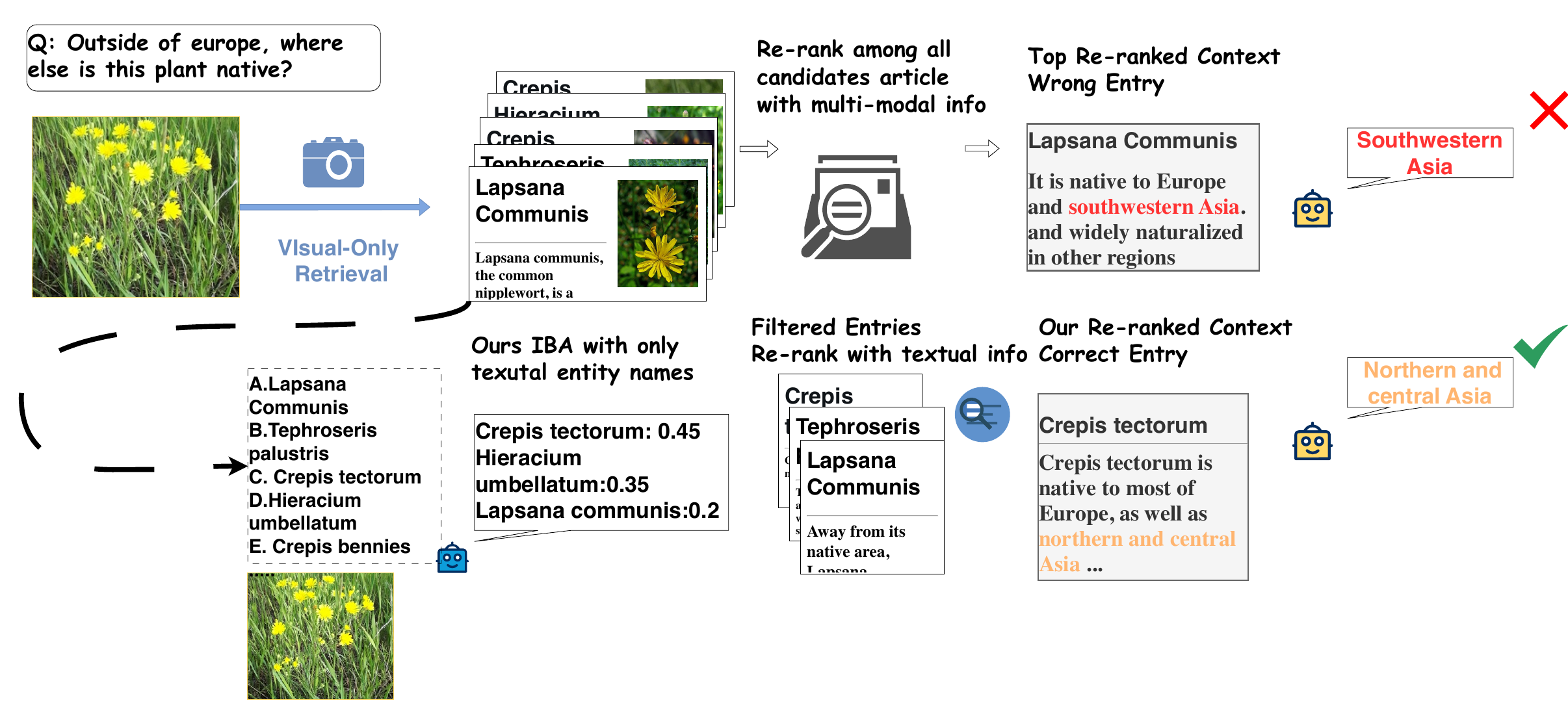}
    \caption{Overall workflow comparison between existing MM-RAG methods and our proposed \ours.
\textbf{Upper}: Existing MM-RAG pipelines retrieve top-$K$ candidate articles from a large knowledge base and directly perform section-level re-ranking using trained re-rankers or fine-tuned VLMs, before generating the answer from the top-ranked context.
\textbf{Lower}: Our training-free \ours inserts an explicit entity identification step before re-ranking.
Given the query image and retrieved candidate names, the VLM selects a small set of high-confidence entities, which are then used to narrow the scope of section-level re-ranking with an off-the-shelf textual re-ranker.
}
\label{fig:KB-VQA-workflow}
\vspace{-3mm}
\end{figure*}

\subsection{Problem Formulation}

Given a query image $I$ and question $Q$, a KB-VQA system aims to generate an answer $y$ by grounding external knowledge.
In retrieval-augmented generation, this is typically achieved by selecting a supporting text snippet from an external knowledge base.

We model the knowledge base as $KB=\{(P_i, I_i)\}_{i=1}^N$, where each page $P_i$ consists of multiple textual sections $P_i=\{S_{i,j}\}_{j=1}^{n_i}$ and is associated with a representative image $I_i$.
Due to the large scale of the knowledge base ($N$ can be millions), practical systems first retrieve a small candidate set and then perform fine-grained re-ranking.
The objective of KB-VQA is to select the most relevant section $S_{i,j}$ that provides sufficient evidence to support generating the correct answer $y$.

\subsection{Initial Retrieval}

The goal of initial retrieval is to obtain a small set of candidate knowledge entries from a massive external knowledge base.
This coarse-grained step ensures tractability by narrowing the search space from millions of entries to a manageable top-$K$ set.

Following prior work~\cite{yan2024echosight,ReflectiVA,CoRe-MMRAG}, we adopt an image-to-image retrieval strategy.
Each knowledge base page is indexed using a frozen EVA-CLIP-8B vision encoder~\cite{sun2024eva}.
Image embeddings are pooled from the final layer and indexed using FAISS~\cite{douze2025faiss}.
Given a query image, cosine similarity is used to retrieve the top-$K$ visually similar candidate pages, which are then passed to subsequent identify-before-re-rank stage.

\subsection{Identify-Before-Re-Rank}

After initial retrieval, existing MM-RAG methods directly perform section-level re-ranking over all $K$ candidate entries.
In contrast, we explicitly introduce an entity-level identification step to further reduce the re-ranking scope.

As we have discussed in Section~\ref{sec:intro}, one of the key challenges in KB-VQA is entity identification to secure grounding at the entity level.
Even for human experts, directly naming the exact real-world entity depicted in an image can be non-trivial, especially when visual cues are subtle or the entity belongs to a fine-grained category.
This difficulty is further amplified for MLLMs under open-ended generation settings, where the model must produce the correct entity name from a vast output space without explicit constraints.

At the same time, modern MLLMs have been trained on large-scale, high-quality corpora that include extensive encyclopedic knowledge, much of which originates from Wikipedia-style resources.
Such training endows models with latent expertise that can support entity recognition, but this expertise is often difficult to reliably elicit through unconstrained generation.
We hypothesize that the challenge lies not in the absence of knowledge, but in the form of the task: open-ended entity naming imposes a high uncertainty burden, whereas selecting the correct entity from a small candidate set is a more tractable discriminative problem.
As empirically validated in Section~\ref{sec:retrieval}, prompting MLLMs to select from candidate entities yields substantially higher identification accuracy than open-ended naming.
This behavior is analogous to the human tip-of-the-tongue phenomenon~\cite{brown1966tip}, where experts may struggle to recall an exact name spontaneously but can readily identify the correct option when presented with a shortlist of plausible candidates.

Motivated by this observation, we design the identification step by prompting the MLLM with a list of candidate entity names, rather than asking it to generate the entity name freely.
This formulation allows the model to focus on assessing relative relevance among plausible candidates, effectively activating its latent encyclopedic knowledge while avoiding the brittleness of open-ended generation.

Accordingly, given the retrieved candidate set $\{(P_i, I_i)\}_{i=1}^K$, we prompt the MLLM with:
(i) query image $I$,
(ii) textual names of all $K$ candidate entries,
and (iii) their initial visual retrieval similarity scores.

The MLLM is asked to assess entity relevance and select the top-$j$ candidates ($j<K$).
This produces an identification confidence score $\mathrm{ID}(P_i)$ for each candidate entry, reflecting the model’s belief that the entity depicted in the image corresponds to $P_i$.

For each identified entry, we compute textual relevance between the question $Q$ and each section $S_{i,j}$ using a pre-trained textual re-ranker such as BGE~\cite{chen2024bge}.
The re-ranker outputs a normalized textual relevance score $\mathrm{T}(S_{i,j}) \in [0,1]$.
Unlike multimodal re-rankers used in prior work~\cite{yan2024echosight}, this component is used off-the-shelf without task-specific training.

We combine identification confidence, visual similarity from initial retrieval, and textual relevance to compute a final score:
\[
\mathrm{score}(S_{i,j}) = \alpha \cdot \mathrm{ID}(P_i) + \beta \cdot \mathrm{V}(I, I_i) + \gamma \cdot \mathrm{T}(S_{i,j}),
\]
where $\alpha$, $\beta$, and $\gamma$ control the contribution of each signal.
For E-VQA, we set $(\alpha,\beta,\gamma)=(0.5,0.5,1)$.
For InfoSeek, we increase $\alpha$ to emphasize entity identification due to weaker visual alignment between query images and knowledge base images.
{The sensitivity analysis in Section~\ref{sec:retrieval} reveals that the final re-ranking outcome is not sensitive to the hyper-parameters combination, since setting all to 1 has modest degradation.}
The top-ranked section is selected as supporting context for answer generation.

Overall, the identify-before-re-rank design provides a simple yet effective alternative to existing MM-RAG pipelines.
By explicitly decoupling entity-level identification from section-level evidence selection, our framework avoids the need for training specialized multimodal re-rankers or fine-tuning large vision--language models.
This decoupling also improves interpretability, as the contributions of visual similarity, entity identification, and textual relevance can be examined independently.
Moreover, restricting section-level scoring to a small set of identified entities substantially reduces computational cost and context length, leading to more efficient inference.
Finally, because our method relies only on off-the-shelf components, it generalizes naturally across different knowledge bases and KB-VQA benchmarks without dataset-specific adaptation.

\subsection{Answer Generation}

Once the top-ranked supporting section is obtained, we use off-the-shelf large language models to generate the final answer.
Our framework does not require fine-tuning the generation model, making it flexible across different backbones.
Compared with prior approaches that rely on fine-tuned multi-modal generators~\cite{ReflectiVA,CoRe-MMRAG}, our method improves answer quality by providing more precise and entity-grounded context, rather than modifying the generation model itself.

\section{Experiments}

\subsection{Datasets and External Knowledge Base}

We evaluate on two KB-VQA benchmarks: Encyclopedic VQA (E-VQA)~\cite{EVQA} and InfoSeek~\cite{InfoSeek}, where answering requires external knowledge beyond the query image.
E-VQA contains 221K image-question pairs (up to five images per question) and covers both single-hop and two-hop questions.
Following prior work~\cite{yan2024echosight}, we focus on the single-hop setting.
Importantly, E-VQA provides a controlled knowledge base of 2M Wikipedia articles with associated images, ensuring that each QA pair is answerable when the correct article is retrieved.

InfoSeek contains 1.3M QA pairs over 11K visual entities from OVEN~\cite{OVEN}.
Following existing MM-RAG baselines~\cite{yan2024echosight,CoRe-MMRAG,ReflectiVA}, we adopt the 100K-article knowledge base released by~\citet{yan2024echosight} and report results on the validation split following the same settings.

\subsection{Evaluation Metrics}

\noindent \textbf{Retrieval.}
We report Recall@$K$, which measures whether the ground-truth article appears in the top-$K$ retrieved candidates.
Following prior work~\cite{yan2024echosight,ReflectiVA}, a retrieved article is counted as correct only if its URL exactly matches the target page URL.
We report Recall@1 as a proxy for \emph{top-1 entity selection accuracy} after re-ranking, reflecting how well a method prioritizes correct entity among retrieved candidates.

\noindent \textbf{Answer generation.}
For E-VQA, we evaluate open-ended answers using the BEM score~\cite{zhang2019bertscore}.
For InfoSeek, we follow prior practice~\cite{yan2024echosight,ReflectiVA} and use VQA accuracy~\cite{goyal2017making,marino2019ok} for time and numerical questions, and BEM score for string questions.

\subsection{Implementation Details}

\noindent \textbf{Retriever and candidate set.}
Following prior work~\cite{yan2024echosight,ReflectiVA}, we use EVA-CLIP-8B~\cite{sun2024eva} for image-to-image retrieval and retrieve the top-$K{=}20$ candidate articles from the knowledge base released by~\citet{yan2024echosight}.

\noindent \textbf{Identifier and Answer generators.}
After initial retrieval, Qwen-2.5-VL-7B-Instruct is deployed to implement the explicit identification step.
We instantiate our pipeline with two off-the-shelf backbones for answer generation: Llama-3.1-8B-Instruct and Qwen-2.5-VL-7B-Instruct (also used for identification).

\noindent \textbf{Baselines.}
For EchoSight~\cite{yan2024echosight}, we follow the original pipeline and apply its released multimodal re-ranker to re-rank sections from the same top-$K$ retrieved candidates, using the default weighting between initial retrieval similarity and re-ranking scores.
For ReflectiVA~\cite{ReflectiVA}, we run the officially released model to produce REL tokens and generate answers by conditioning on sections assigned REL tokens within the top-5 retrieved articles.

\noindent \textbf{Zero-shot variants.}
Following the two-stage prompting design in Core-MMRAG~\cite{CoRe-MMRAG}, we implement several zero-shot variants to probe the role of workflow design.
Given the top-5 retrieved articles, \emph{1Stage} prompts the MLLM to directly answer with all articles as context, while \emph{2Stage} first selects the most relevant article and then generates the answer conditioned on that article only.
We also include \emph{Para} (no external evidence) and \emph{Article} (directly use the top-5 retrieved articles without explicit re-ranking) variants.

\subsection{Retrieval Results}\label{sec:retrieval}

The retrieval results on InfoSeek~\cite{InfoSeek} and E-VQA~\cite{EVQA} are reported in Tables~\ref{tab:retrieval_infoseek} and~\ref{tab:retrieval_evqa}.
EVA-CLIP retrieval yields moderate Recall@20 but much lower Recall@1, showing that while the correct entity is usually present among candidates, it is rarely ranked first by visual similarity alone (e.g., E-VQA Recall@1: 13.4\%, Recall@20: 48.8\%).
This highlights the need for a re-ranking stage to better prioritize the correct entity.

After applying re-ranking, both EchoSight and our proposed \ours substantially improve Recall@1, confirming the importance of re-ranking for entity prioritization.
On InfoSeek, our method outperforms EchoSight by a clear margin, achieving Recall@1 of 58.4\% compared to 53.1\% (Table~\ref{tab:retrieval_infoseek}).
This improvement indicates that the explicit identify-before-re-rank design is more effective at prioritizing the correct entity from visually similar candidates.
\begin{table}[!htbp]
\caption{InfoSeek retrieval results.
EVA-CLIP denotes the initial image-to-image retrieval using EVA-CLIP-8B.
EchoSight applies its trained multimodal re-ranker on top of the same retrieved candidates.
Our method prompts the MLLM to select the top-3 entities from the 20 retrieved candidate names.
}
\centering
\begin{tabular}{@{}cccccc@{}}
\toprule
\multirow{2}{*}{Method} & \multicolumn{5}{c}{InfoSeek Recall@k}              \\ \cmidrule(l){2-6} 
                        & k=1           & k=3           & k=5  & k=10 & k=20 \\ \midrule
EVA-CLIP                & 45.6          & 63.1          & 68.6 & 74.6 & 77.9 \\
EchoSight               & 53.1          & 69.4          & 73.9 & 77.4 & 77.9 \\
Our \ours                    & \textbf{58.4} & \textbf{72.4} & -    & -    & -   \\ \bottomrule
\end{tabular}
\label{tab:retrieval_infoseek}
\vspace{-3mm}
\end{table}

On E-VQA, EchoSight achieves slightly higher Recall@1 (36.5\%) than our method (35.5\%), as shown in Table~\ref{tab:retrieval_evqa}.
This outcome is expected, as EchoSight is specifically trained on E-VQA using curated positive supervision.
In contrast, our approach is entirely training-free and directly transferable across datasets.
Despite this small gap in Recall@1, the downstream generation results (Section~\ref{sec:gen}) show that higher identification accuracy alone does not guarantee better answer quality.
\begin{table}[!htbp]
\caption{E-VQA retrieval results.
EVA-CLIP denotes the initial image-to-image retrieval using EVA-CLIP-8B.
EchoSight applies its trained multimodal re-ranker on top of the retrieved candidates.
Our method prompts the MLLM to select the top-3 entities from the 20 retrieved candidate names.
}
\centering
\begin{tabular}{@{}cccccc@{}}
\toprule
\multirow{2}{*}{Method} & \multicolumn{5}{c}{E-VQA Recall@k}                 \\ \cmidrule(l){2-6} 
                        & k=1           & k=3           & k=5  & k=10 & k=20 \\ \midrule
EVA-CLIP                & 13.4          & 26.1          & 31.9 & 41.8 & 48.8 \\
EchoSight               & \textbf{36.5} & \textbf{45.3} & 47.9 & 48.8 & 48.8 \\
Our \ours                    & 35.5          & 43.3          & -    & -    & -    \\ \bottomrule
\end{tabular}
\label{tab:retrieval_evqa}
\vspace{-3mm}
\end{table}

\noindent \textbf{Grounded-subset identification analysis.}
To directly probe entity identification (independent of answer generation), we evaluate on grounded subsets where each question is guaranteed to be answerable from its ground-truth entity page.
On E-VQA, our method correctly identifies the ground-truth entity for 934/2,322 grounded questions (40.2\%), compared to 593/2,322 (25.5\%) under open-ended entity naming.
On InfoSeek, we randomly sample 1,000 validation questions, among which 790 are grounded; our method identifies correctly for 578/790 (73.2\%) versus 362/790 (45.8\%) under open-ended naming.
These results further support providing candidate entity names substantially amplifies MLLM-based identification.
To demonstrate that this phenomenon is not specific to a single model, we further tested an advanced proprietary model, GPT-5.2~\cite{openai_gpt52_systemcard_2025}. 
On the grounded subset of E-VQA, GPT-5.2’s identification ratio improved from 23.4\% to 58.0\% by providing textual options.

\noindent \textbf{Sensitivity Analysis}.
We conduct a small sensitivity analysis by setting all score-fusion weights (ID score, visual similarity, text relevance) to 1, removing dataset-specific tuning.
Under this setting, Recall@1 is 34.9\% on 
E-VQA (vs. 35.5\%) and 56.3\% on InfoSeek (vs. 58.4\%). 
The drops are modest ($ - 0.6\%$ and $ - 2.1\%$), indicating limited sensitivity to weight choices.
Even without tuning, IBA remains competitive with EchoSight~\cite{yan2024echosight} (53.1\% on InfoSeek), whose re-ranker requires supervised training. This suggests that the improvement is largely attributable to the workflow design rather than hand-tuned weight optimization.

\subsection{Generation Results}
\label{sec:gen}

We evaluate answer generation quality on InfoSeek and E-VQA and compare our training-free pipeline with finetuned MM-RAG baselines and zero-shot variants (Table~\ref{tab:gen_main}).

\begin{table*}[!htbp]
\small
\centering
\caption{Answer generation results on E-VQA and InfoSeek.} 
\begin{tabular}{llllllllll}
\toprule
\multicolumn{1}{c}{}                          & \multicolumn{1}{c}{}                           & \multicolumn{7}{c}{InfoSeek}                                                                                                                                                                                      & \multicolumn{1}{c}{}                        \\ \cmidrule(lr){3-9}
\multicolumn{1}{c}{}                          & \multicolumn{1}{c}{}                           & \multicolumn{1}{c}{}                          & \multicolumn{3}{c}{Unseen Question}                                             & \multicolumn{3}{c}{Unseen Entity}                                               & \multicolumn{1}{c}{}                        \\ \cmidrule(lr){4-9}
\multicolumn{1}{c}{\multirow{-3}{*}{Methods}} & \multicolumn{1}{c}{\multirow{-3}{*}{Backbone}} & \multicolumn{1}{c}{\multirow{-2}{*}{Overall}} & \multicolumn{1}{c}{time} & \multicolumn{1}{c}{num} & \multicolumn{1}{c}{string} & \multicolumn{1}{c}{time} & \multicolumn{1}{c}{num} & \multicolumn{1}{c}{string} & \multicolumn{1}{c}{\multirow{-3}{*}{E-VQA}} \\ \midrule
\rowcolor{cyan!15}\multicolumn{10}{l}{\textit{Our proposed \ours}} \\
\ours-Qwen                                     & Qwen-2.5-VL                                    & {\ul 37.2}                                    & 34.5                     & 7.8                     & 47.5                       & 40.1                     & 6.6                     & 45.2                       & \textbf{43.6}                               \\
\ours-LLaVA                                    & Llama-3.1-8B                                   & \textbf{37.8}                                 & 38.7                     & 13.5                    & 45.9                       & 44.5                     & 12.8                    & 44.4                       & {\ul 43.2}                                  \\
\rowcolor{yellow}\multicolumn{10}{l}{\textit{Retrieve Augmented Models requiring Fine-tuning}}                                                                                                                                                                                                                                                                                    \\
ReflectiVA                                    & Llama-3.1-8B                                   & 36.4                                          & 29.7                     & 10.4                    & 45.6                       & 36.5                     & 12.1                    & 43.2                       & 38.6                                        \\
EchoSight                                     & Llama-3.1-8B                                   & 33.8                                          & 24.9                     & 12.3                    & 41.7                       & 37.5                     & 11.3                    & 39.8                       & 41.9                                        \\
\rowcolor{lightgray} \multicolumn{10}{l}{\textit{Zero-shot base models}}                                                                                                                                                                                                                                                                                                           \\
Para-Llava                                    & Llama-3.1-8B                                   & 9.0                                           & 0.9                      & 0.5                     & 12.4                       & 2.4                      & 0.5                     & 10.2                       & 13.3                                        \\
Para-Qwen                                     & Qwen-2.5-VL                                    & 25.5                                          & 10.9                     & 0.0                     & 35.5                       & 12.1                     & 0.0                     & 32.9                       & 21.2                                        \\
\rowcolor{lightgray} \multicolumn{10}{l}{\textit{Zero-shot with Retrieval}}                                                                                                                                                                                                                                                                                      \\
Article-llava                                 & Llama-3.1-8B                                   & 18.4                                          & 0.0                      & 0.0                     & 26.4                       & 0.0                      & 0.0                     & 24.1                       & 23.0                                        \\
Article-qwen                                  & Qwen-2.5-VL                                    & 34.5                                          & 14.3                     & 0.2                     & 46.8                       & 12.1                     & 0.5                     & 45.9                       & 35.6                                        \\
\rowcolor{lightgray} \multicolumn{10}{l}{\textit{Zero-shot with Re-rank}}                                                                                                                                                                                                                                                                                                   \\
1Stage-llava                                  & Llama-3.1-8B                                   & 10.5                                          & 0.0                      & 0.0                     & 15.0                       & 0.0                      & 0.0                     & 13.9                       & 4.0                                         \\
1Stage-Qwen                                   & Qwen-2.5-VL                                    & 34.6                                          & 32.7                     & 1.5                     & 44.8                       & 40.1                     & 2.1                     & 44.4                       & 34.3                                        \\
2Stage-llava                                  & Llama-3.1-8B                                   & 27.2                                          & 6.9                      & 0.9                     & 38.5                       & 5.0                      & 0.2                     & 36.4                       & 23.1                                        \\
2Stage-Qwen                                   & Qwen-2.5-VL                                    & 35.9                                          & 8.4                      & 0.0                     & 49.0                       & 4.5                      & 0.3                     & 48.5                       & 34.1                                        \\ \bottomrule
\end{tabular}
\label{tab:gen_main}
\end{table*}


\noindent \textbf{RAG vs.\ purely parametric MLLMs.}
Across both backbones, retrieve-augmented methods substantially outperform purely parametric variants (\emph{Para-*}), highlighting that external evidence is essential for KB-VQA.
On InfoSeek, \emph{Para-LLaVA} and \emph{Para-Qwen} achieve 9.0 and 25.5 overall, while retrieve-augmented variants reach 30--38.
On E-VQA, \emph{Para-LLaVA} and \emph{Para-Qwen} obtain 13.3 and 21.2, and our proposed \ours exceed 43.

\noindent \textbf{Training-free identify-before-answer vs.\ finetuned MM-RAG.}
Our training-free pipeline outperforms finetuned baselines on both datasets.
On InfoSeek, our \emph{\ours-LLaVA} attains the best overall score (37.8), followed by \emph{\ours-Qwen} (37.2), both surpassing ReflectiVA (36.4) and EchoSight (33.8).
On E-VQA, \emph{\ours-Qwen} and \emph{\ours-LLaVA} achieve 43.6 and 43.2, outperforming EchoSight (41.9) and ReflectiVA (38.6).
Notably, our method requires no task-specific training or additional parameters, while EchoSight and ReflectiVA rely on finetuned components.
A closer look at InfoSeek question types shows that our improvements are consistent on time questions, where correct entity grounding and evidence selection are critical.
For example, on unseen-question time queries, \emph{\ours-LLaVA} achieves 38.7 versus 29.7 (ReflectiVA) and 24.9 (EchoSight), and on unseen-entity time queries it achieves 44.5 versus 36.5 and 37.5, respectively.
More qualitative results are shown in Appendix~\ref{app:qual}.

\subsection{Ablation Study}

\textbf{Zero-shot MLLM.}
To test whether an off-the-shelf MLLM can solve KB-VQA by prompting alone, we follow~\citet{CoRe-MMRAG} and compare four zero-shot variants that differ in how retrieved evidence is used.
\emph{Para-*} answers using only parametric knowledge (no external evidence).
\emph{Article-*} answers in one step given the full text of the top-5 retrieved articles.
\emph{1Stage-*} further adds the retrieved entry images and explicitly asks the model to use the most relevant reference, effectively requiring implicit multimodal re-ranking inside a single long prompt.
\emph{2Stage-*} decomposes this into two prompts: the model first selects the most relevant article and then generates answer conditioned on that article only.

\noindent\textbf{External evidence helps, but prompting is not a reliable re-ranker.}
Across both datasets, moving from \emph{Para-*} to \emph{Article-*} yields large gains, confirming that KB-VQA cannot be solved reliably from parametric knowledge alone.
For LLaVA, InfoSeek improves from 9.0 (\emph{Para-LLaVA}) to 18.4 (\emph{Article-LLaVA}), and E-VQA improves from 13.3 to 23.0.
For Qwen, InfoSeek improves from 25.5 (\emph{Para-Qwen}) to 34.5 (\emph{Article-Qwen}), and E-VQA improves from 21.2 to 35.6.
However, asking the model to implicitly re-rank evidence within a single prompt is unstable: \emph{1Stage-LLaVA} collapses to 10.5 on InfoSeek and 4.0 on E-VQA, despite using more information than \emph{Article-LLaVA}.

\noindent\textbf{Two-stage prompting helps, but still trails a structured pipeline.}
Decomposing the interaction partially mitigates the above issue: \emph{2Stage-LLaVA} recovers to 27.2 on InfoSeek and 23.1 on E-VQA, yet remains far behind our full pipeline (37.8 and 43.2, respectively).
For Qwen, explicit re-ranking provides limited or inconsistent gains: \emph{1Stage-Qwen} and \emph{2Stage-Qwen} reach 34.6/35.9 on InfoSeek (vs.\ 34.5 for \emph{Article-Qwen}) and slightly underperform on E-VQA (34.3/34.1 vs.\ 35.6).
Overall, these results suggest that while strong backbones can exploit long retrieved text, prompting alone does not reliably perform evidence selection, motivating an explicit and lightweight identification-and-reranking workflow.

\begin{figure*}[!htbp]
\begin{minipage}{\linewidth}
{\textbf{Q}: What does this fish eat?}\\
\begin{minipage}{0.3\linewidth}
\includegraphics[width=0.7\linewidth]{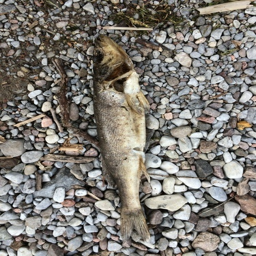}
\end{minipage}
\hfill
\begin{minipage}{0.7\linewidth}
\tiny{

\textbf{EchoSight~\cite{yan2024echosight}}:\\
Walleye is a nocturnal feeder, which means it feeds at night. It is most easily caught using live minnows or lures that mimic small fish. This implies that the walleye's \textcolor{red}{diet consists of small fish, likely minnows.} \textcolor{red}{\xmark} \\
{\#\# SELECTED Section Title: As food \\
Walleye is considered to be a palatable freshwater fish, consequently, is fished recreationally and commercially for food.  Because of its nocturnal feeding habits, it is most easily caught \textcolor{red}{at night using live minnows or lures that mimic small fish} ... \\}
\textbf{\ours (Ours):}\\
Walleyes eat a variety of food sources depending on their age and development. Here are specific food sources mentioned:\textcolor{orange}{* Invertebrates, such as fly larvæ and zooplankton, when the young walleyes are absorbing their yolk sac.* Fish, such as yellow perch or ciscoes, when they become piscivorous after 40-60 days.* Crayfish, minnows, and leeches, in addition to fish, when they become adults.} \textcolor[HTML]{00b050}{\cmark} \\
{\#\# SELECTED Section Title: Reproduction \\
 ... Once the yolk has been fully absorbed, the young walleyes \textcolor{orange}{begin to feed on invertebrates, such as fly larvæ and zooplankton.} After 40 to 60 days, juvenile walleyes become piscivorous. \textcolor{orange}{Thenceforth, both juvenile and adult walleyes eat fish almost exclusively, frequently yellow perch or ciscoes, moving onto bars and shoals at night to feed.} ...
 }
}
\end{minipage}
\end{minipage}

\caption{Qualitative result E-VQA, where we compare the answers provided by \ours with EchoSight~\cite{yan2024echosight} that more informative context is selected by our \ours.}
\label{fig:informative-context-main}
\vspace{-0.5cm}
\end{figure*}
\noindent \textbf{EchoSight's multimodal re-ranker vs.\ our proposed re-rank after identification.}
To disentangle the effect of entity identification from section selection, we evaluate several re-ranking variants on the E-VQA grounded subset, where the correct entity page is guaranteed to contain the answer.
We compare (i) EchoSight's original pipeline, (ii) our full identify-before-answer model, and two hybrids that keep our identification module but replace our textual re-ranker with EchoSight's released multimodal re-ranker, either without (\emph{\ours - MIS}) or with (\emph{\ours + MIS}) incorporating the MLLM identification score (MIS) into the final ranking.
Table~\ref{tab:ablation} summarizes the results.

\begin{table}[!htbp]
\caption{Ablation on E-VQA grounded questions comparing re-ranking choices after entity identification.
\ours -/+ MIS replace our textual re-ranker with EchoSight's released multimodal re-ranker, without /with using the MLLM identification score (MIS) in final ranking.
}
\small
\centering
\begin{tabular}{@{}lcc@{}}
\toprule
\multicolumn{1}{c}{Methods}                            & Identified Ratio & Score \\ \midrule
EchoSight                                              & 77.8             & 66.4 \\
\ours $-$ MIS w EchoSight reranker                         & 73.6             & 63.2 \\
\ours $+$ MIS w EchoSight reranker & 75.1             & 62.2 \\
\ours full                                            & 75.7             & 70.5 \\ \bottomrule
\end{tabular}
\label{tab:ablation}
\vspace{-3mm}
\end{table}

We draw two observations from Table~\ref{tab:ablation}.
First, replacing our textual re-ranker with EchoSight's multi-modal re-ranker consistently reduces answer quality, even when the identified ratio is comparable (around 73--78\%).
For example, EchoSight and the two hybrid variants obtain scores of 66.4/63.2/62.2, all below our method (70.5).
Second, incorporating MIS does not improve the hybrid: \emph{\ours + MIS} slightly increases the identified ratio over \emph{\ours -- MIS} (75.1 vs.\ 73.6) but further lowers the final score (62.2 vs.\ 63.2).
These results suggest that, once the entity is (mostly) correct, effective KB-VQA hinges on selecting \emph{informative} sections rather than visually plausible but weak evidence.
To further analyze this effect, Table~\ref{tab:breakdown_evqa} breaks down performance on grounded questions by whether EchoSight and our method identify the ground-truth entity.

\begin{table}[!htbp]
\small
\caption{Breakdown results on E-VQA grounded questions. Our \ours achieves better overall results mainly from selecting more informative sections than EchoSight. Some samples are shown in Figure~\ref{fig:informative-context-main}~and~\ref{fig:informative-context}.}
\begin{tabular}{@{}cccccc@{}}
\toprule
            & \thead{Both \textcolor[HTML]{00b050}{\cmark} \\ (59.3\%)} & \thead{Our \textcolor[HTML]{00b050}{\cmark} \\ (13.4\%)} & \thead{Echo \textcolor[HTML]{00b050}{\cmark} \\ (15.5\%)} & \thead{Both 
            \textcolor{red}{\xmark} \\ (11.8\%)} & Overall \\ \midrule
Echo & 79.6               & 22.8               & 80.6                 & 30.9                   & 66.4 \\
\ours  & 88.2               & 83.6                & 23.9                 & 27.6                   & 70.5
 \\ \bottomrule
\end{tabular}
\label{tab:breakdown_evqa}
\end{table}

Although EchoSight attains a slightly higher identified ratio (77.8\% vs.\ 75.7\%), our method achieves a higher overall answer score (70.5 vs.\ 66.4).
When both methods identify the correct entity (59.3\% of questions), our score is substantially higher (88.2 vs.\ 79.6), indicating more informative section selection given the same entity.
Our advantage is even more pronounced when only our method identifies the correct entity (13.4\%): we maintain a strong score (83.6) while EchoSight often fails to provide useful evidence (22.8).
Conversely, EchoSight performs better only when it identifies the correct entity and we do not (15.5\% of questions; 80.6 vs.\ 23.9).
When both miss the entity (11.8\%), both methods perform poorly.

To reflect the evidence quality in a more straightforward way, we further check the direct evidence hits. For the automatically generated subset of E-VQA (2,750 questions), where evidence annotations are available, we check Evidence Hit (if the evidence selected by method exactly matches the annotated evidence) in addition to Entity Matching as shown in Table~\ref{tab:evidence-hit}.

\begin{table}[!htbp]
\centering
\caption{ Direct evidence hit results on E-VQA automatically generated subset in percentage. While the entity match is lower than EchoSight~\cite{yan2024echosight}, our proposed IBA achieves higher direct evidence hit ratio.
}
\label{tab:evidence-hit}
\begin{tabular}{@{}lcc@{}}
\toprule
Method    & Entity Match & Evidence Hit \\ \midrule
IBA       & 37.1         & 33.8         \\
EchoSight & 39.3         & 30.2         \\ \bottomrule
\end{tabular}
\end{table}

While entity match is lower, IBA achieves higher Evidence Hit. 
This pattern aligns with our decomposition analysis in Table~\ref{tab:breakdown_evqa}, where we show that correct entity identification alone does not guarantee correct answers without proper evidence grounding.
Together, these results indicate that the performance gain of IBA does not stem solely from entity recall, but from improved evidence-level grounding enabled by explicitly decoupling identification from ranking.

Overall, these ablations indicate that our gains are not solely due to entity identification.
Explicitly decoupling identification from purely textual re-ranking leads to more reliable selection of supportive evidence.
Even when the same entity is retrieved, our approach tends to choose sections that directly contain the facts required by the question, whereas a trained multimodal re-ranker can favor visually plausible but less informative sections.
Qualitative examples in Figures~\ref{fig:informative-context-main} and~\ref{fig:informative-context} further illustrate this difference.

\section{Conclusion}

In this work, we revisit KB-VQA from a workflow perspective and identify entity and evidence level groundings as critical bottlenecks.
While recent MLLMs possess substantial encyclopedic knowledge, we show that this knowledge is difficult to reliably elicit under open-ended entity naming.
Instead, MLLMs exhibit significantly stronger identification ability when selecting from a small set of candidate entities.
Motivated by this observation, we propose a simple and training-free identify-before-answer framework that explicitly decouples entity identification from section-level evidence selection.
By prompting MLLMs with candidate entity names and leveraging an off-the-shelf textual re-ranker for evidence selection, our approach avoids the need for specialized multi-modal re-ranker training while remaining interpretable and robust.
Extensive experiments on E-VQA and InfoSeek demonstrate that this decomposition consistently outperforms finetuned MM-RAG baselines.
Our analyses also reveal that effective KB-VQA depends not only on identifying the correct entity, but also on selecting informative supporting evidence once the entity is fixed.
We hope this work encourages future research to rethink retrieval-augmented reasoning pipelines by explicitly separating distinct grounding and selection stages, and to explore more lightweight designs for knowledge-intensive multi-modal reasoning.
\section*{Acknowledgements}
This research is supported by the National Science Foundation (NSF) under grant numbers NSF-2406647 and NSF-2406648.
It is also supported by the National Artificial Intelligence Research Resource (NAIRR) Pilot and the Delta advanced computing and data resource, which is supported by the National Science Foundation under award NSF-OAC-2005572.
\section*{Limitations}
Although our proposed \ours surpasses existing fine-tuning baselines and demonstrates impressive performance on Knowledge-based VQA like Encyclopedic-VQA and InfoSeek,
we note the following limitation that there is a dependence on the external knowledge base.
In real-world application, it could be possible that the knowledge base is not perfect to include all supporting evidence for answering questions.
However, there are emerging works focused on integrating agentic workflow into the VQA tasks~\cite{jiang2024mmsearch,wu2025mmsearch,fu2025livevqa} at the retrieval stage. Specifically, they will invoke external search tools \ie the whole Internet will be considered as the external knowledge base, which could be a promising direction for future works.

\bibliography{main}

\appendix
\section{Additional Related Work}

\subsection{Multimodal Large Language Models}

Multimodal large language models (MLLMs) extend text-only LLMs by integrating modality encoders and alignment mechanisms, enabling joint reasoning over images and text.
Early models such as Flamingo~\cite{alayrac2022flamingo} and BLIP-2~\cite{li2023blip} demonstrated that coupling a pretrained vision encoder with a frozen language model can already yield strong few-shot multimodal capabilities.
Subsequent models, including the LLaVA family~\cite{liu2023visual,liu2024improved} and Qwen-VL~\cite{bai2025qwen2}, further advanced this paradigm through large-scale instruction tuning and improved cross-modal fusion architectures, achieving strong performance across a wide range of multimodal benchmarks.

Despite these advances, recent empirical studies~\cite{li2024cognitive,tan2025vision} consistently show that even state-of-the-art MLLMs underperform on knowledge-based VQA (KB-VQA) benchmarks that require fine-grained, entity-centric, or long-tail factual knowledge.
This limitation motivates augmenting MLLMs with external knowledge sources to support explicit grounding and reasoning beyond parametric knowledge alone.

\subsection{Knowledge-Based Visual Question Answering}

Conventional visual question answering (VQA) benchmarks~\cite{antol2015vqa,goyal2017making,wang2017fvqa} focus on questions that can be answered using visual content and common-sense knowledge.
With large-scale pretraining and instruction tuning, modern MLLMs perform competitively on these tasks, as much of the required information is either directly observable or implicitly encoded during training~\cite{yin2024survey,bai2025qwen2}.

Knowledge-based VQA (KB-VQA) fundamentally extends this setting by requiring external knowledge not contained in the image alone, such as entity attributes or encyclopedic facts~\cite{deng2025comprehensive}.
Early benchmarks, including OK-VQA~\cite{marino2019ok} and A-OKVQA~\cite{schwenk2022okvqa}, introduced questions that depend on commonsense or general knowledge, which increasingly falls within the training scope of large-scale MLLMs.

More recent benchmarks, such as Encyclopedic-VQA~\cite{EVQA} and InfoSeek~\cite{InfoSeek}, further raise the difficulty by requiring fine-grained, entity-level grounding in Wikipedia.
These datasets demand explicit identification of the correct entity and selection of supporting sections from retrieved articles.
Despite progress in MLLMs, models continue to struggle in this setting due to challenges in discriminating visually similar entities and selecting the most informative evidence, motivating retrieval-augmented approaches.

\subsection{MM-RAG-Based Solutions}

To address the limitations of parametric knowledge in KB-VQA, many recent methods adopt multimodal retrieval-augmented generation (MM-RAG).
As illustrated in Figure~\ref{fig:KB-VQA-workflow}, MM-RAG pipelines typically consist of three stages: (i) a retriever that performs coarse retrieval from a large-scale knowledge base, (ii) a re-ranking stage that selects the most relevant context among retrieved candidates, and (iii) an answer generator that produces the final response conditioned on the selected evidence.

Representative methods differ primarily in how relevance is modeled.
EchoSight~\cite{yan2024echosight} trains a dedicated multimodal re-ranker to select the most relevant sections given an image–question pair, leveraging contrastive learning with curated supervision.
However, obtaining high-quality positive supervision for section-level relevance is challenging, and among major KB-VQA benchmarks, only E-VQA provides explicit supporting section annotations.

Other approaches, such as ReflectiVA~\cite{ReflectiVA} and CoRe-MMRAG~\cite{CoRe-MMRAG}, rely on fine-tuning large MLLMs to implicitly assess relevance through internal representations.
While effective, these methods require constructing task-specific training data and incur substantial computational cost when fine-tuning large-scale models such as LLaVA~\cite{liu2024improved} or Qwen-VL~\cite{bai2025qwen2}.
These limitations motivate alternative designs that reduce training overhead while improving robustness and transferability.

\section{Qualitative results}\label{app:qual}
\begin{figure*}[t]
\begin{minipage}{0.325\linewidth}
\scriptsize{infoseek-00016131 \textbf{Q}: In which year was this equipment retired from operational service? \\ \textbf{Query Entity: Bren light machine gun} \vspace{0.05cm}}\\
\begin{minipage}{0.443\linewidth}
\includegraphics[width=1.\linewidth]{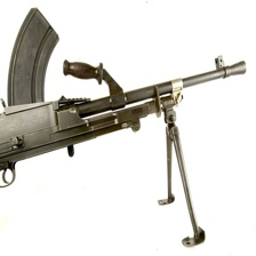}
\end{minipage}
\hfill
\begin{minipage}{0.53\linewidth}
\scriptsize{
\textbf{ReflectiVA~\cite{ReflectiVA}}:\\
1963 \textcolor{red}{\xmark} \\
\textbf{EchoSight~\cite{yan2024echosight}}:\\
Not mentioned in the text. \textcolor{red}{\xmark} \\
\textbf{\ours (Ours):}\\
2006 \textcolor[HTML]{00b050}{\cmark}
}
\end{minipage}
\end{minipage}
\hspace{0.02cm}
\begin{minipage}{0.325\linewidth}
\scriptsize{00073156 \textbf{Q}:What is the sea level in metre of this mountain? \\ \textbf{Query Entity:Corcovado} \vspace{0.05cm}}\\
\begin{minipage}{0.443\linewidth}
\includegraphics[width=1.\linewidth]{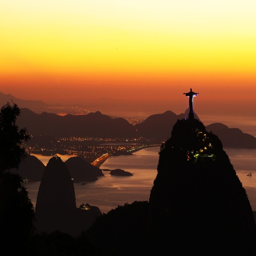}
\end{minipage}
\hfill
\begin{minipage}{0.53\linewidth}
\scriptsize{
\textbf{ReflectiVA~\cite{ReflectiVA}}:\\
396 \textcolor{red}{\xmark} \\
\textbf{EchoSight~\cite{yan2024echosight}}:\\
396 \textcolor{red}{\xmark} \\
\textbf{\ours (Ours):}\\
710 \textcolor[HTML]{00b050}{\cmark}
}
\end{minipage}
\end{minipage}
\hspace{0.02cm}
\begin{minipage}{0.325\linewidth}
\scriptsize{00073009\textbf{Q}: what was the date this aircraft enter into service? \\ \textbf{Query Entity:Tupolev Tu-154} \vspace{0.05cm}}\\
\begin{minipage}{0.443\linewidth}
\includegraphics[width=1.\linewidth]{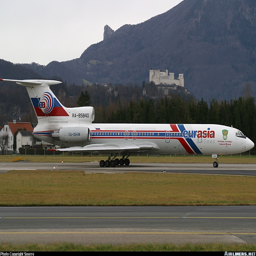}
\end{minipage}
\hfill
\begin{minipage}{0.53\linewidth}
\scriptsize{
\textbf{ReflectiVA~\cite{ReflectiVA}}:\\
1970 2 21 \textcolor{red}{\xmark} \\
\textbf{EchoSight~\cite{yan2024echosight}}:\\
Not mentioned. \textcolor{red}{\xmark} \\
\textbf{\ours (Ours):}\\
9 February 1972 \textcolor[HTML]{00b050}{\cmark}
}
\end{minipage}
\end{minipage}
\vspace{0.1cm}

\begin{minipage}{0.325\linewidth}
\scriptsize{00072497 \textbf{Q}: In which year did this building come into service? \\ \textbf{Query Entity:Niechorze Lighthouse 1866} \vspace{0.05cm}}\\
\begin{minipage}{0.443\linewidth}
\includegraphics[width=1.\linewidth]{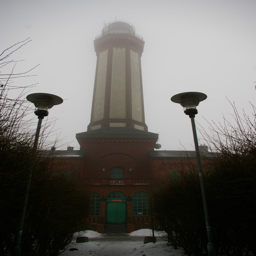}
\end{minipage}
\hfill
\begin{minipage}{0.53\linewidth}
\scriptsize{
\textbf{ReflectiVA~\cite{ReflectiVA}}:\\
1814 \textcolor{red}{\xmark} \\
\textbf{EchoSight~\cite{yan2024echosight}}:\\
1814 \textcolor{red}{\xmark} \\
\textbf{\ours (Ours):}\\
1866 \textcolor[HTML]{00b050}{\cmark}
}
\end{minipage}
\end{minipage}
\hspace{0.02cm}
\begin{minipage}{0.325\linewidth}
\scriptsize{00073353 \textbf{Q}: What is the length of this lake in kilometre?\\ \textbf{Query Entity:Windermere} \vspace{0.05cm}}\\
\begin{minipage}{0.443\linewidth}
\includegraphics[width=1.\linewidth]{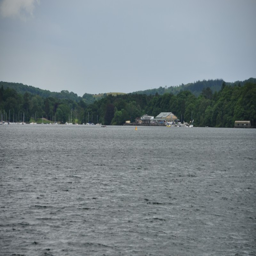}
\end{minipage}
\hfill
\begin{minipage}{0.53\linewidth}
\scriptsize{
\textbf{ReflectiVA~\cite{ReflectiVA}}:\\
11.23 \textcolor{red}{\xmark} \\
\textbf{EchoSight~\cite{yan2024echosight}}:\\
11 \textcolor{red}{\xmark} \\
\textbf{\ours (Ours):}\\
18 \textcolor[HTML]{00b050}{\cmark}
}
\end{minipage}
\end{minipage}
\hspace{0.02cm}
\begin{minipage}{0.325\linewidth}
\scriptsize{00072995 \textbf{Q:} What is the weight of a male of this bird in gram? \\ \textbf{Query Entity:Least grebe 129} \vspace{0.05cm}}\\
\begin{minipage}{0.443\linewidth}
\includegraphics[width=1.\linewidth]{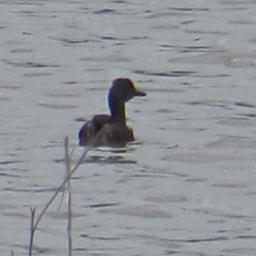}
\end{minipage}
\hfill
\begin{minipage}{0.53\linewidth}
\scriptsize{
\textbf{ReflectiVA~\cite{ReflectiVA}}:\\
692 \textcolor{red}{\xmark} \\
\textbf{EchoSight~\cite{yan2024echosight}}:\\
692-925 \textcolor{red}{\xmark} \\
\textbf{\ours (Ours):}\\
129 \textcolor[HTML]{00b050}{\cmark}
}
\end{minipage}
\end{minipage}

\vspace{0.1cm}

\begin{minipage}{0.325\linewidth}
\scriptsize{00018660 \textbf{Q}: What is the closest upper taxonomy of this bird? \\ \textbf{Query Entity:Eurasian collared dove} \vspace{0.05cm}}\\
\begin{minipage}{0.443\linewidth}
\includegraphics[width=1.\linewidth]{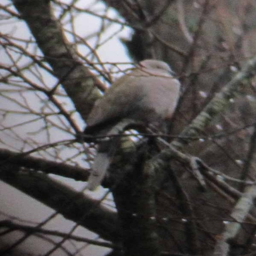}
\end{minipage}
\hfill
\begin{minipage}{0.53\linewidth}
\scriptsize{
\textbf{ReflectiVA~\cite{ReflectiVA}}:\\
Zenaida \textcolor{red}{\xmark} \\
\textbf{EchoSight~\cite{yan2024echosight}}:\\
Genus \textcolor{red}{\xmark} \\
\textbf{\ours (Ours):}\\
Streptopelia \textcolor[HTML]{00b050}{\cmark}
}
\end{minipage}
\end{minipage}
\hspace{0.02cm}
\begin{minipage}{0.325\linewidth}
\scriptsize{00018687 \textbf{Q}: What is this plant named after? \\ \textbf{Query Entity:Allamanda} \vspace{0.05cm}}\\
\begin{minipage}{0.443\linewidth}
\includegraphics[width=1.\linewidth]{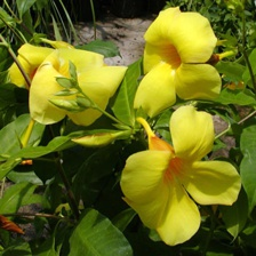}
\end{minipage}
\hfill
\begin{minipage}{0.53\linewidth}
\scriptsize{
\textbf{ReflectiVA~\cite{ReflectiVA}}:\\
Charles Plumier \textcolor{red}{\xmark} \\
\textbf{EchoSight~\cite{yan2024echosight}}:\\
Thevetia \textcolor{red}{\xmark} \\
\textbf{\ours (Ours):}\\
Frédéric-Louis Allamand \textcolor[HTML]{00b050}{\cmark}
}
\end{minipage}
\end{minipage}
\hspace{0.02cm}
\begin{minipage}{0.325\linewidth}
\scriptsize{00000129 \textbf{Q:} What country does this building belong to? \\ \textbf{Query Entity:Polish Baltic Philharmonic} \vspace{0.05cm}}\\
\begin{minipage}{0.443\linewidth}
\includegraphics[width=1.\linewidth]{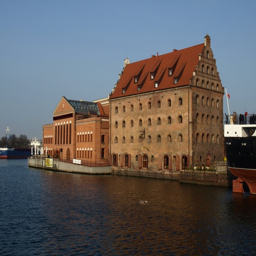}
\end{minipage}
\hfill
\begin{minipage}{0.53\linewidth}
\scriptsize{
\textbf{ReflectiVA~\cite{ReflectiVA}}:\\
Netherlands \textcolor{red}{\xmark} \\
\textbf{EchoSight~\cite{yan2024echosight}}:\\
Denmark \textcolor{red}{\xmark} \\
\textbf{\ours (Ours):}\\
Poland \textcolor[HTML]{00b050}{\cmark}
}
\end{minipage}
\end{minipage}

\vspace{-0.2cm}
\caption{Sample qualitative results on image-question pairs from InfoSeek, where we compare the answers provided by \ours with those from ReflectiVA~\cite{ReflectiVA} and EchoSight~\cite{yan2024echosight}.}
\label{fig:qualiatives}
\vspace{-0.14cm}
\end{figure*}

\begin{figure*}[t]
\begin{minipage}{0.325\linewidth}
\scriptsize{\textbf{Q}: How long is this bird in centimeters? \\ \textbf{Query Entity: Sanderling} \vspace{0.05cm}}\\
\begin{minipage}{0.443\linewidth}
\includegraphics[width=1.\linewidth]{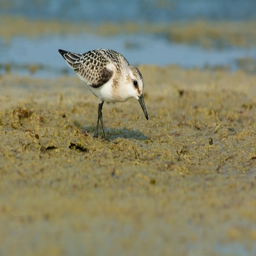}
\end{minipage}
\hfill
\begin{minipage}{0.53\linewidth}
\scriptsize{
\textbf{ReflectiVA~\cite{ReflectiVA}}:\\
30–40 \textcolor{red}{\xmark} \\
\textbf{EchoSight~\cite{yan2024echosight}}:\\
The breeding adult bird is 14-16 cm in length. \textcolor{red}{\xmark} \\
\textbf{\ours (Ours):}\\
The sanderling is 18–20 cm (7.1–7.9 in) in length. \textcolor[HTML]{00b050}{\cmark}
}
\end{minipage}
\end{minipage}
\hspace{0.02cm}
\begin{minipage}{0.325\linewidth}
\scriptsize{\textbf{Q}: Who designed this hotel? \\ \textbf{Query Entity:The Fullerton Hotel Singapore} \vspace{0.05cm}}\\
\begin{minipage}{0.443\linewidth}
\includegraphics[width=1.\linewidth]{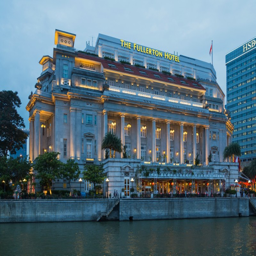}
\end{minipage}
\hfill
\begin{minipage}{0.53\linewidth}
\scriptsize{
\textbf{ReflectiVA~\cite{ReflectiVA}}:\\
 The hotel rooms were designed by Hirsch Bedner Associates. \textcolor{red}{\xmark} \\
\textbf{EchoSight~\cite{yan2024echosight}}:\\
 Hirsch Bedner Associates designed the hotel rooms. \textcolor{red}{\xmark} \\
\textbf{\ours (Ours):}\\
The Fullerton Building was designed by Major P.H. Keys of Keys \& Dowdeswell, a Shanghai firm of architects. \textcolor[HTML]{00b050}{\cmark}
}
\end{minipage}
\end{minipage}
\hspace{0.02cm}
\begin{minipage}{0.325\linewidth}
\scriptsize{\textbf{Q}: What settlements comprised the this abbey estate? \\ \textbf{Query Entity: Bolton Abbey} \vspace{0.05cm}}\\
\begin{minipage}{0.443\linewidth}
\includegraphics[width=1.\linewidth]{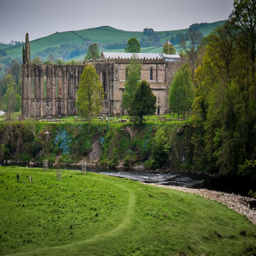}
\end{minipage}
\hfill
\begin{minipage}{0.53\linewidth}
\scriptsize{
\textbf{ReflectiVA~\cite{ReflectiVA}}:\\
villages \textcolor{red}{\xmark} \\
\textbf{EchoSight~\cite{yan2024echosight}}:\\
\textcolor{red}{The Fountains Abbey} estate comprised the following settlements:... \textcolor{red}{\xmark} \\
\textbf{\ours (Ours):}\\
The Bolton Abbey estate comprised the following settlements: Addingham,Holme,Beamsley and Gargrave \textcolor[HTML]{00b050}{\cmark}
}
\end{minipage}
\end{minipage}
\vspace{0.1cm}

\begin{minipage}{0.325\linewidth}
\scriptsize{\textbf{Q}: What tool is associated with the traditional seaside atmosphere of this pier? \\ \textbf{Query Entity:Brighton Palace Pier} \vspace{0.05cm}}\\
\begin{minipage}{0.443\linewidth}
\includegraphics[width=1.\linewidth]{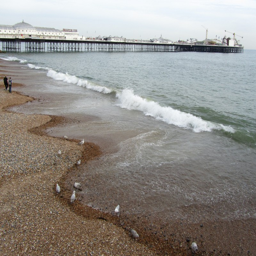}
\end{minipage}
\hfill
\begin{minipage}{0.53\linewidth}
\scriptsize{
\textbf{ReflectiVA~\cite{ReflectiVA}}:\\
whelk \textcolor{red}{\xmark} \\
\textbf{EchoSight~\cite{yan2024echosight}}:\\
The answer is: Carousel. \textcolor{red}{\xmark} \\
\textbf{\ours (Ours):}\\
The traditional seaside atmosphere of the Brighton Palace Pier is associated with ""bucket and spade"". \textcolor[HTML]{00b050}{\cmark}
}
\end{minipage}
\end{minipage}
\hspace{0.02cm}
\begin{minipage}{0.325\linewidth}
\scriptsize{\textbf{Q}: What is the architectural style of this mosque? \\ \textbf{Query Entity: Great Mosque of Banten} \vspace{0.05cm}}\\
\begin{minipage}{0.443\linewidth}
\includegraphics[width=1.\linewidth]{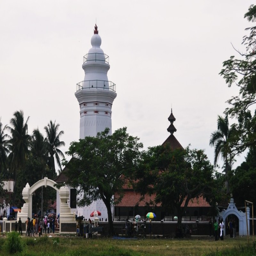}
\end{minipage}
\hfill
\begin{minipage}{0.53\linewidth}
\scriptsize{
\textbf{ReflectiVA~\cite{ReflectiVA}}:\\
Islamic architecture \textcolor{red}{\xmark} \\
\textbf{EchoSight~\cite{yan2024echosight}}:\\
The architectural style of the Great Mosque of Banten is a mix of Indian Mughal pattern and ancient candi decoration. \textcolor{red}{\xmark} \\
\textbf{\ours (Ours):}\\
The architectural style of the Great Mosque of Banten is described as follows joglo style, a traditional Javanese style \textcolor[HTML]{00b050}{\cmark}
}
\end{minipage}
\end{minipage}
\hspace{0.02cm}
\begin{minipage}{0.325\linewidth}
\scriptsize{\textbf{Q:} When was this park created? \\ \textbf{Query Entity: Princes Street Gardens} \vspace{0.05cm}}\\
\begin{minipage}{0.443\linewidth}
\includegraphics[width=1.\linewidth]{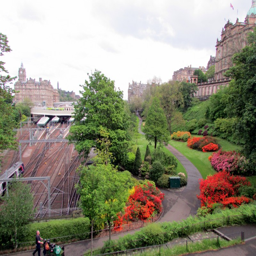}
\end{minipage}
\hfill
\begin{minipage}{0.53\linewidth}
\scriptsize{
\textbf{ReflectiVA~\cite{ReflectiVA}}:\\
1843 \textcolor{red}{\xmark} \\
\textbf{EchoSight~\cite{yan2024echosight}}:\\
The park was created in 1876, when the town council reacquired the ground and laid it out as a public park. \textcolor{red}{\xmark} \\
\textbf{\ours (Ours):}\\
Princes Street Gardens were created in the 1820s. \textcolor[HTML]{00b050}{\cmark}
}
\end{minipage}
\end{minipage}

\vspace{-0.2cm}
\caption{Sample qualitative results on image-question pairs from E-VQA, where we compare the answers provided by \ours with those from ReflectiVA~\cite{ReflectiVA} and EchoSight~\cite{yan2024echosight}.}
\label{fig:qualiatives-evqa}
\vspace{-0.14cm}
\end{figure*}

In this section we will provide more qualitative results on both InfoSeek and E-VQA datasets as shown in Figure~\ref{fig:qualiatives} and~\ref{fig:qualiatives-evqa}.

\section{More examples for informative context}
\begin{figure*}[!htbp]
\begin{minipage}{\linewidth}
{\textbf{Q}: How many tiny reddish yellow warts \\ does each segment of this moth have? \\ \textbf{Query Entity: Atolmis \\ rubricollis} \vspace{ -0.05cm}}\\
\begin{minipage}{0.3\linewidth}
\includegraphics[width=0.9\linewidth]{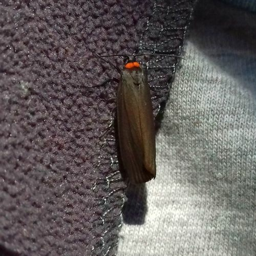}
\end{minipage}
\hfill
\begin{minipage}{0.7\linewidth}
\scriptsize{

\textbf{EchoSight~\cite{yan2024echosight}}:\\
According to the text, \textcolor{red}{there is no mention} of tiny reddish-yellow warts on the segments of the red-necked footman moth. The description only mentions the coloration of the thorax, head, abdomen, antennae, legs, and wings, but does not mention any warts. Therefore, the answer is None. \textcolor{red}{\xmark} \\
\#\# Section Title: Description\\
The red-necked footman is a small moth that is mostly charcoal grey or deep dark brown (fresh specimens almost black), but has a conspicuous orange thorax, part of which is visible behind the black head as an orange-red collar. The hindwings are brownish grey. The antennae and legs are black and the end of the abdomen is yellowish orange or golden yellow. The wings are tightly folded together around the body and have pleated, squared-off ends. The wingspan is 25 to 35 mm (1.0 to 1.4 in) and the length of the forewings is 15 to 18 mm (0.6 to 0.7 in).
Wilhelm Neuburger named Romanian specimens as the synonym Atolmis flavicollis; these were less dark brown and with a light yellow instead of a red collar. \\
\textbf{\ours (Ours):}\\
According to the text, \textcolor{orange}{each segment of the caterpillar bears six tiny reddish yellow warts.} \textcolor[HTML]{00b050}{\cmark} \\
\#\# Section Title: Life cycle\\
The white eggs of the red-necked footman are laid in small groups in crevices in the branches of trees, especially those of old firs. They grow to a length of about 27 mm (1.1 in). Their head is black with a bold diagonal white stripe on either side. Their main colour is dark greenish grey marbled with cream. \textcolor{orange}{Each segment bears six tiny reddish yellow warts which bear black hairs.} The caterpillars feed on lichens growing on the trunks and branches of trees, and can be found between August and October...
}
\end{minipage}
\end{minipage}
\vspace{5mm}

\begin{minipage}{\linewidth}
{\textbf{Q}: In which year(s) was this castle rebuilt? \\ \textbf{Query Entity: Dudley Castle} \vspace{0.05cm}}\\
\begin{minipage}{0.3\linewidth}
\includegraphics[width=0.9\linewidth]{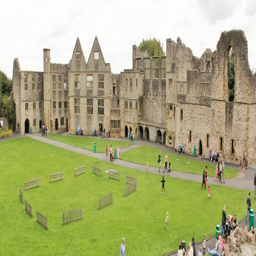}
\end{minipage}
\hfill
\begin{minipage}{0.7\linewidth}
\scriptsize{

\textbf{EchoSight~\cite{yan2024echosight}}:\\
The answer is: None. According to the text, the bulk of the remaining habitable parts of the castle was \textcolor{red}{destroyed by fire in 1750}, and there is \textcolor{red}{no mention of the castle being rebuilt.} \textcolor{red}{\xmark} \\
SELECTED Section Title: Final years and ruin\\ 
stable block was constructed on the site at some point before 1700. This was the final building to be constructed in the castle.
The bulk of the remaining habitable parts of the castle was \textcolor{red}{destroyed by fire in 1750}. However, in the nineteenth century, the site found a new use as a 'Romantic Ruin' and a certain amount of tidying up of the site was carried out by the Earls of Dudley. Battlements on one of the remaining towers were reconstructed and two cannon captured during the Crimean Wars were installed. In the nineteenth and early twentieth century the site was used for fêtes and pageants. In 1937, when the Dudley Zoo was established, the castle grounds were incorporated into the zoo. ... \\
\textbf{\ours (Ours):}\\
According to the evidence, the castle keep was rebuilt starting from 1262. \textcolor[HTML]{00b050}{\cmark} \\
SELECTED Section Title: The keep\\ 
The castle keep dates \textcolor{orange}{from the rebuilding that started in 1262}. It rests on the motte, constructed in the Norman period but somewhat reduced in height afterwards. The original building was slightly rectangular in plan with approximate dimensions 15 metres north to south and 22 metres east to west. The four drum towers on each corner are 9.8 metres in diameter. After the slighting at the end of the civil war, only the north side of the castle and parts of two of the drum towers remain.
}
\end{minipage}
\end{minipage}

\vspace{5mm}

\begin{minipage}{\linewidth}
{\textbf{Q}:What kind of scabious is this plant commonly known as? \\ \textbf{Query Entity: Knautia Arvensis} \vspace{ -0.01cm}}\\
\begin{minipage}{0.3\linewidth}
\includegraphics[width=0.9\linewidth]{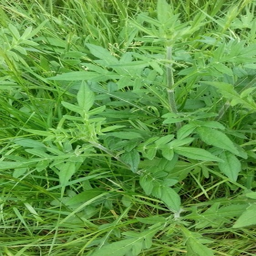}
\end{minipage}
\hfill
\begin{minipage}{0.7\linewidth}
{

\textbf{EchoSight~\cite{yan2024echosight}}:\\
According to the article, the plant is commonly known as \textcolor{red}{Gipsy Rose}. \textcolor{red}{\xmark} \\
SELECTED Section Title: Name\\ 
Species of scabious were used to treat scabies, and many other afflictions of the skin including sores caused by the bubonic plague. The word scabies comes from the Latin word scabere 'scratch'. \textcolor{red}{Another name for this plant is gipsy rose.}
The genus Knautia is named after a 17th-century German botanist, Christian Knaut. \\
\textbf{\ours (Ours):}\\
Based on the evidence text, the answer is: field scabious. \textcolor[HTML]{00b050}{\cmark} \\
SELECTED Section Title: Knautia arvensis\\ 
Knautia arvensis, commonly known as \textcolor{red}{field scabious}, is a herbaceous perennial species of flowering plant in the honeysuckle family Caprifoliaceae.
}
\end{minipage}
\end{minipage}

\vspace{-0.2cm}
\caption{Sample qualitative results on image-question pairs from E-VQA, where we compare the answers provided by \ours with EchoSight~\cite{yan2024echosight} where we selected more informative context.}
\label{fig:informative-context}
\vspace{-0.14cm}
\end{figure*}

We have presented more examples on Figure~\ref{fig:informative-context} where our proposed \ours selects more informative context compared with EchoSight~\cite{yan2024echosight}.
\section{Error Case Study}
\begin{figure*}[!htbp]
\begin{minipage}{\linewidth}
{\textbf{Q}: In which part of the world does this animal live? \\ \textbf{Query Entity: Echinometra Mathaei \\ rubricollis} \vspace{ -0.05cm}}\\
\begin{minipage}{0.3\linewidth}
\includegraphics[width=0.9\linewidth]{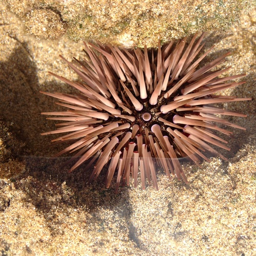}
\end{minipage}
\hfill
\begin{minipage}{0.7\linewidth}
{
\textbf{ReflectiVA~\cite{ReflectiVA}}:\\
World Ocean \\

\textbf{EchoSight~\cite{yan2024echosight}}:\\
The \textcolor{red}{red sea urchin} (Mesocentrotus franciscanus) is found in the northeastern Pacific Ocean, specifically in the areas of Alaska to Baja California. \textcolor{red}{\xmark} \\

\textbf{\ours (Ours):}\\
The \textcolor{red}{purple sea urchin}, Strongylocentrotus purpuratus, lives along the eastern edge of the Pacific Ocean, extending from Ensenada, Mexico, to British Columbia, Canada. \textcolor{red}{\xmark} \\

}
\end{minipage}
\end{minipage}
\vspace{5mm}

\begin{minipage}{\linewidth}
{\textbf{Q}: How big is an adult of this species typically?  \\ \textbf{Query Entity: Argiope bruennichi} \vspace{0.05cm}}\\
\begin{minipage}{0.3\linewidth}
\includegraphics[width=0.9\linewidth]{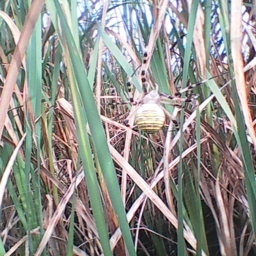}
\end{minipage}
\hfill
\begin{minipage}{0.7\linewidth}
{
\textbf{ReflectiVA~\cite{ReflectiVA}}:\\
insects \\

\textbf{EchoSight~\cite{yan2024echosight}}:\\
The answer is not explicitly stated in the provided text. It does not provide information about the size of the adult \textcolor{red}{Argiope spider}. \textcolor{red}{\xmark} \\

\textbf{\ours (Ours):}\\
I did not find any information about the size of an adult \textcolor{red}{Argiope species}. \textcolor[HTML]{00b050}{\cmark} \\

}
\end{minipage}
\end{minipage}

\vspace{5mm}

\begin{minipage}{\linewidth}
{\textbf{Q}:What did president bill clinton sign to create this memorial? \\ \textbf{Query Entity: Oklahoma City National Memorial. Ground Truth Answer: Law} \vspace{ -0.01cm}}\\
\begin{minipage}{0.3\linewidth}
\includegraphics[width=0.9\linewidth]{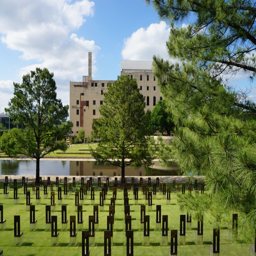}
\end{minipage}
\hfill
\begin{minipage}{0.7\linewidth}
{
\textbf{ReflectiVA~\cite{ReflectiVA}}:\\
Bill \textcolor{red}{\xmark} \\
\textbf{EchoSight~\cite{yan2024echosight}}:\\
The answer is: the Oklahoma City National Memorial Act of 1997. \textcolor{red}{\xmark} \\
\textbf{\ours (Ours):}\\
President Bill Clinton signed the ""Oklahoma City National Memorial Act of 1997"" to create the Oklahoma City National Memorial. \textcolor{red}{\xmark}
}
\end{minipage}
\end{minipage}

\vspace{-0.2cm}
\caption{Error cases on image-question pairs from E-VQA, where we compare the answers provided by our proposed \ours with ReflectiVA~\cite{ReflectiVA} and EchoSight~\cite{yan2024echosight}.}
\label{fig:erros-cases}
\vspace{-0.14cm}
\end{figure*}

In this section, we provide more examples of error cases.
As shown in Figure~\ref{fig:erros-cases}, for most cases, the error is caused by compromised entity grounding as shown in first two cases. 
However, it could be possible that the data annotation is not concise enough or evaluation is not comprehensive enough. 
As shown in the last case, both our proposed \ours and EchoSight have secured the correct entity grounding and provided reasonable answer \ie the exact name of the law signed ` Oklahoma City National Memorial Act '.
However, it is judged as wrong via existing evaluation protocols, which could be caused due to the too coarse-grained or vague ground truth answer \textbf{law.}
\section{Prompt Template}

\subsection{Identification}
We explicitly prompt the MLLM to perform entity identification as a constrained selection task, rather than open-ended entity naming. Given a query image and a small set of retrieved candidate entity names, the model is asked to select the most likely entity (or top-$k$ entities) depicted in the image. 
We also utilize the initial visual retrieval similarity score in the prompts.
The identification prompt is formatted as follows:
\lstset{basicstyle=\ttfamily\footnotesize, breaklines=true, breakindent=0pt}
\begin{lstlisting}
SYSTEM: You are an expert visual entity recognizer. Look at the image and here are some potentially relevant options.

Options:
A. <ENTITY_NAME_1> (image similarity: <SIM_1>)
B. <ENTITY_NAME_2> (image similarity: <SIM_2>)
C. <ENTITY_NAME_3> (image similarity: <SIM_3>)
...

Reply with 'Answer: <label1>, <label2>, ...' listing the top <K> option letters from most to least likely based on the image.
\end{lstlisting}

Here, each option corresponds to a candidate entity retrieved from the external knowledge base, optionally augmented with its initial image-to-image retrieval similarity score. The model is required to respond strictly in the prescribed format, enabling us to directly interpret the output as entity-level confidence for subsequent re-ranking.

\subsection{Answer Generation}
For the answer generation stage, we follow existing works~\cite{yan2024echosight} to apply answer generation templates depending on the dataset.

\paragraph{E-VQA.}
The prompt we use for LLMs when testing Encyclopedic-VQA (E-VQA)~\cite{EVQA} is shown as follows:
\lstset{basicstyle=\ttfamily\footnotesize}
\begin{lstlisting}
USER: Context: <CONTEXT>
Question: <QUESTION>
The answer is:
\end{lstlisting}

\paragraph{InfoSeek.}
Due to the strict exact-match evaluation used by InfoSeek~\cite{InfoSeek}, following existing works~\cite{yan2024echosight}, we adopt a one-shot prompting strategy and add instructions to ensure the generated answer strictly matches the required format. The prompt used for InfoSeek is:
\lstset{basicstyle=\ttfamily\footnotesize, breaklines=true, breakindent=0pt}
\begin{lstlisting}
SYSTEM: You always answer the question the user asks. Do not answer anything else.

USER:Context: The southern side of the Alps is next to Lake Como.
Question: Which body of water is this mountain located in or next to?
Just answer the questions, no explanations needed. 
Short answer is: Lake Como

Context: <CONTEXT>
Question: <QUESTION>
Just answer the questions, no explanations needed.
Short answer is:
\end{lstlisting}
\section{Dataset Details}
Following existing works~\cite{yan2024echosight,CoRe-MMRAG,ReflectiVA}, we use the same test set of InfoSeek \cite{InfoSeek} and E-VQA~\cite{EVQA}, consists of 71,335 and 4,750 question pairs respectively,
\section{Token Budget and Re-ranking Efficiency}
\label{app:rerank-efficiency}
In this section we will introduce more details to compare the cost of our proposed method with existing methods in terms of token and FLOPs.
\subsection{Our proposed \ours with BGE Reranker}
The Qwen top-$k$ pipeline operates as follows:
\begin{enumerate}
  \item Retrieve top 20 entities per query; Qwen identification keeps the top 3 entities.
  \item For the retained entities, obtain all wiki sections and send them to the BGE section reranker.
  \item Obtain the section scores by comparing all sections with the query text to select the best section for downstream answer generation.
\end{enumerate}
Empirically, our prepared metadata for E-VQA shows:
\begin{align*}
  \text{avg. sections/example} &\approx 24.7,\\
  \text{avg. tokens/section}   &\approx 172.
\end{align*}
Assuming questions contribute $\approx 20$ tokens, the total BGE token budget per example is
\[
  24.7 \times (172 + 20) \approx 4.8\text{k tokens},
\]
.

\subsection{EchoSight Re-ranker}
The EchoSight flow differs from our proposed \ours with larger re-ranking space.
\begin{enumerate}
  \item Retrieve top 20 entities, which is same as ours.
  \item Expand all entities to sections, encode the image once, and encode \emph{all} sections with BLIP-2 Q-Former.
  \item Score by query(both image and text) and candidate sections relevance(similarity) and rerank.
\end{enumerate}
Using the same section statistics as a proxy, the section pool grows to
\[
  20 \times \tfrac{24.7}{3} \approx 165 \text{ sections},
\]
yielding a text load of $\approx 165 \times 172 \approx 28\text{k tokens}$, i.e., about $6\!-\!7\times$ more text than our proposed \ours.

\subsection{Runtime Implications (full pipeline)}
\begin{itemize}
  \item \textbf{Ours}
    \begin{itemize}
      \item Identification (Qwen2.5-VL-7B): $\approx$256 visual tokens (one image encode) + $70$--$100$ text tokens input (question + 20 candidate titles) $\Rightarrow$ $400$ tokens.
      \item Section rerank (BGE, text-only): $\approx 4.8\text{k}$ text tokens (24.7 sections $\times$ (172 per section + 20 for the question)).
    \end{itemize}
    Overall cost is dominated by the BGE text forward; the image is encoded once in identification.
  \item \textbf{EchoSight reranker (no identification):}
    \begin{itemize}
      \item One image encode ($\approx$256 visual tokens).
      \item Text side encodes around $28\text{k}$ text tokens ($\approx 165$ sections $\times$ 172 tokens per section) and fuses with the image. The model(\ie EchoSight Re-ranker based on BLIP-2) is heavier than the off-the-shelf textual re-ranker such as bge-v2-m3.
    \end{itemize}
\end{itemize}

Thus, end-to-end, our identification + text re-rank load is far smaller than EchoSight’s multi-modal re-rank; the gains come from entity pruning (section count reduced $6\!-\!7\times$) and using a lightweight text re-ranker.

\section{Computational Cost Analysis via FLOPs}
\label{appendix:flops}

We analyze computational cost using floating-point operations (FLOPs) rather than wall-clock latency.
Wall-clock time is highly sensitive to implementation details, hardware configuration, batching strategy, and system-level optimizations, making it difficult to fairly compare methods with different architectures.
In contrast, FLOPs provide a hardware-agnostic and reproducible proxy for inference complexity that reflects the intrinsic computational demand of a model.
This metric has been widely adopted in prior work for comparing model efficiency across architectures and modalities.

\paragraph{FLOPs estimation protocol.}
To compare inference cost across different model architectures in a hardware-agnostic and reproducible manner, we estimate computational complexity using floating-point operations (FLOPs). Following prior work that analyzes the scaling and efficiency of Transformer models~\cite{kaplan2020scaling}, we adopt a standard proxy for Transformer-based modules:

For a Transformer encoder with hidden dimension $d$ and $L$ layers, the forward FLOPs per token can be approximated as
\begin{equation}
\text{FLOPs per token} \approx 24 \times L \times d^2,
\end{equation}
which captures the dominant contributions of multi-head self-attention and feed-forward network operations across layers. The total compute is then obtained by multiplying this per-token cost by the number of tokens processed by the model. Using this unified protocol for all Transformer-only components (e.g., transformer layers in text or cross-modal encoders) ensures a fair basis for comparison across methods.

Vision Transformers (ViT)~\cite{dosovitskiy2020vit}, however, require a different consideration because their self-attention operations compute pairwise interactions among image patch tokens, resulting in computation that scales quadratically with sequence length. Specifically, self-attention involves operations on $S\times S$ affinity matrices, where $S$ is the number of visual tokens, dominating compute when $S$ is large. 
For these vision encoders, we therefore estimate FLOPs by accounting explicitly for both the attention term ($O(S^2 \cdot d)$) and the feed-forward term ($O(S \cdot d \cdot d_{ff})$), rather than relying solely on the $d^2$ proxy. This exception reflects the inherent quadratic complexity of self-attention in visual processing and is consistent with practice in ViT analysis~\cite{touvron2022three,marin2023token,zhang2023mg}.

\paragraph{EchoSight.}
EchoSight performs multi-modal re-ranking using a BLIP-2-based~\cite{li2023blip} architecture.
During re-ranking, EchoSight does not invoke a large language model.
Instead, it employs the BLIP-2 Querying Transformer (Q-Former), a BERT~\cite{devlin2019bert} style Transformer (12 layers, hidden size 768), to encode both the multi-modal query (image + question) and all candidate wiki sections, where vision information will be processed with its Vision Transformer~\cite{dosovitskiy2020vit} vision encoder \footnote{\url{https://github.com/Go2Heart/EchoSight/blob/main/lavis/models/blip2_models/blip2_qformer_reranker.py}}
.
Each candidate section is encoded independently using the same Q-Former in a text-only mode, and relevance scores are computed via embedding similarity.
Given that EchoSight expands all retrieved entities into sections, the re-ranking stage processes a large number of text tokens, leading to substantial computational cost.
We estimate the FLOPs of EchoSight re-ranking by accounting for:
(i) the frozen vision encoder,
(ii) the Q-Former multi-modal fusion, and
(iii) the Q-Former-based text encoding of all candidate sections.
For the vision encoder, EchoSight uses an EVA-CLIP Vision Transformer variant with a patch size of 14, hidden dimension $d_v=1408$, and depth $L_v=39$ transformer layers. The input image of size $224\times224$ is tokenized into $16^2=256$ patches plus one classification token, yielding $S=257$ tokens.

To estimate FLOPs for Vision Transformers (ViT), we decompose the dominant contributions as:
\begin{itemize}
  \item \textbf{Self-attention:} computing the attention score matrix $QK^T$ involves $O(S^2 \cdot d)$ operations due to pairwise interactions among tokens.
  \item \textbf{Feed-forward network (FFN):} each token is transformed via two linear layers with intermediate dimension $d_{ff}\approx4d$, contributing $O(S\cdot d \cdot d_{ff})$ operations.
\end{itemize}
This yields the per-layer FLOPs approximation:
\[
\text{FLOPs per layer}_{\text{ViT}} \approx 2\,d_v\,S^2\;+\;4\,d_v\,d_{ff}\,S,
\]
where the first term corresponds to self-attention and the second to FFN. Such decomposition reflects the quadratic dependence on token count intrinsic to self-attention in vision models. 

Substituting $S=257$, $d_v=1408$, $L_v=39$ and $d_{ff}\approx4d_v$, the FLOPs for a single forward pass through the vision encoder can be approximated as:
\begin{align}
\text{FLOPs}_{vision} 
&\approx L_v\!\left(2\,d_v\,S^2 + 4\,d_v\,d_{ff}\,S\right) \nonumber\\
& \approx 3.3\times10^{11}.
\end{align}

This indicates that a single forward pass through EVA-CLIP’s Vision Transformer backbone incurs on the order of $10^{11}$ FLOPs, consistent with the understanding that self-attention costs grow quadratically with the number of tokens processed.

For the multi-modal fusion and text encoding, EchoSight employs the BLIP-2 Querying Transformer (Q-Former)~\cite{li2023blip}, a BERT-base style Transformer with hidden size $d_q=768$ and $L_q=12$ layers. Using the same transformer FLOPs proxy yields the following cost per token.,
\begin{align}
    \nonumber & \text{FLOPs}_{QFormer} \approx 24\times L_q \times d_q^2 \\
    & \approx 24\times12\times768^2\approx1.7\times10^{10}.
\end{align}

For the multi-modal fusion step, this cost is incurred once for the query and question tokens.
For section text encoding, each candidate section is encoded with the same Q-Former in text-only mode. Given an approximate rerank pool of $\sim28{,}000$ tokens, the total FLOPs for text encoding becomes
\begin{equation}
28{,}000\times1.7\times10^{10}\;\approx\;4.8\times10^{14}.
\end{equation}

Overall, while the vision encoder and fusion contribute on the order of $10^{10}$–$10^{11}$ FLOPs, the text encoding FLOPs dominate the re-ranking cost at $\sim4.8\times10^{14}$. These estimates use consistent transformer FLOPs proxies and model configuration data from the EchoSight implementation, supporting a concrete and fair comparison of computational demand across methods.

\paragraph{Our proposed \ours FLOPs.}
For our pipeline, the computational cost comprises two stages: (i) multi-modal identification using Qwen2.5-VL-7B Instruct and (ii) text-only re-ranking using the BAAI/bge-reranker-v2-m3 model.

\textbf{Identification (Qwen2.5-VL-7B Instruct).}
Qwen2.5-VL-7B Instruct adopts a re-engineered vision–language architecture rather than a standard ViT encoder followed by a text-only transformer~\cite{bai2025qwen2}. In particular, the vision encoder is redesigned with windowed attention and multi-stage token merging, which significantly reduces the effective sequence length compared to a naive patch-based Vision Transformer. As a result, the quadratic $S^2$ self-attention cost characteristic of vanilla ViT is largely mitigated in practice.

For FLOPs estimation, we therefore treat the identification stage as a unified transformer-style forward pass and apply the standard transformer FLOPs proxy to the shared multi-modal backbone. According to the official configuration, Qwen2.5-VL-7B uses hidden size $d_{\text{Qwen}}=3584$ and $L_{\text{Qwen}}=28$ transformer layers~\cite{bai2025qwen2}. The per-token FLOPs is estimated as
\begin{align}
\nonumber & \text{FLOPs}_{\text{Qwen2.5-VL}} \approx 24 \times L_{\text{Qwen}} \times d_{\text{Qwen}}^2 \\
& \approx 24 \times 28 \times 3584^2 \approx 8.62\times10^{9}.
\end{align}

During identification, the model processes one image together with the question and candidate entity names. Due to the internal vision token compression in Qwen2.5-VL, the effective number of tokens participating in the transformer layers is substantially smaller than the raw patch count. We conservatively approximate the overall forward pass as involving $\sim400$ effective tokens, yielding
\begin{equation}
\text{FLOPs}_{\text{identification}} \approx 400 \times 8.62\times10^{9} \approx 3.45\times10^{12}.
\end{equation}

This estimate intentionally over-approximates the identification cost and provides a conservative upper bound for comparison.

\textbf{Section re-ranking (bge-reranker-v2-m3).}
The off-the-shelf textual re-ranker(bge-reranker-v2-m3)\cite{chen2024bge} uses a RoBERTa-based\cite{liu2019roberta} cross-encoder with hidden size $d_{\text{BGE}}=1024$ and $L_{\text{BGE}}=24$ layers. Again applying the transformer FLOPs proxy yields the following per token cost,
\begin{align}
\nonumber & \text{FLOPs}_{\text{BGE}} \approx 24\times L_{\text{BGE}}\times d_{\text{BGE}}^2 \\
& \approx 24\times24\times1024^2 \approx 6.04\times10^{8}.
\end{align}
Given a rerank pool of $\sim4{,}800$ text tokens per example, the total re-ranking FLOPs becomes
\begin{equation}
\text{FLOPs}_{\text{BGE rerank}} \approx 4800\times6.04\times10^{8} \approx 2.90\times10^{12}.
\end{equation}

Combining both stages gives
\begin{equation*}
\text{FLOPs}_{\text{ours}} \approx 3.45\times10^{12} + 2.90\times10^{12} \approx 6.35\times10^{12},
\end{equation*}
indicating that our pipeline’s compute proxy is dominated by identification and re-ranking FLOPs, and is orders of magnitude lower than the comparable EchoSight re-ranking cost.

\paragraph{Key efficiency advantage.}
A key source of efficiency in our pipeline stems from architectural decoupling, which reduces compute by \emph{nearly two orders of magnitude} compared to EchoSight multi-modal re-ranker.

Under the same FLOPs proxy, EchoSight’s re-ranking cost is dominated by the cost of encoding the entire candidate section pool with a high-capacity multi-modal encoder. Specifically, using an EVA-CLIP Vision Transformer backbone ($4\times10^{10}$ FLOPs per image) and a BLIP-2 Q-Former ($1.7\times10^{10}$ FLOPs per token), the re-ranking stage on a pool of $\sim28{,}000$ text tokens incurs $\sim4.8\times10^{14}$ FLOPs. By contrast, our method incurs only $\sim3.45\times10^{12}$ FLOPs for the multi-modal identification stage with Qwen2.5-VL-7B and $\sim2.90\times10^{12}$ FLOPs for the BGE reranker text stage, totaling $6.35\times10^{12}$ FLOPs. 

This means that even without considering the one-off vision encoding cost, our rerank-centric compute is \emph{nearly two orders of magnitude lower} than the text encoding cost alone in EchoSight’s re-ranker. The bulk of EchoSight’s compute arises from repeatedly processing all candidate sections with a multi-modal model, whereas our approach confines expensive multi-modal processing to a single identification pass and relies on a lightweight text re-ranker for large-scale comparison. This structural decoupling yields \textbf{significantly lower computational demand} while preserving re-ranking effectiveness, demonstrating that effective knowledge-based VQA does not require heavy multi-modal encoding over the entire candidate space.

\end{document}